\documentclass[a4paper, 10pt, conference]{ieeeconf}      

\IEEEoverridecommandlockouts                              

\usepackage{graphics} 
\usepackage{amsmath} 
\usepackage{amssymb}  

\usepackage{cite}
\usepackage{tikz}
\usepackage{xcolor}
\usepackage{tkz-euclide}
\usetikzlibrary{calc,patterns,angles,quotes,patterns,patterns.meta}
\usepackage{pgfplots}
\usepackage{mathtools}
\usepackage{caption}
\usepackage{subcaption}
\usepackage{hyperref}
\usepackage{ifthen}
\newboolean{showFigures}
\setboolean{showFigures}{true}
\newboolean{use_tikz_instead_of_pdf}
\setboolean{use_tikz_instead_of_pdf}{false}
\DeclareMathOperator*{\signum}{sig} 
\newcommand{\matvar}[1]{\boldsymbol{#1}}
\newcommand{\refEquation}[1]{(\ref{#1})}

\newcommand{\refSection}[1]{Sec.~\ref{#1}}

\newcommand{\refFigure}[1]{Fig.~\ref{#1}}

\usepackage{tikz}
\usepackage{textcomp}
\usepackage{hyperref}
\usepackage{lipsum}
\newcommand\copyrighttext{%
  \footnotesize \textcopyright 2025 IEEE. Personal use of this material is permitted.
  Permission from IEEE must be obtained for all other uses, in any current or future
  media, including reprinting/republishing this material for advertising or promotional
  purposes, creating new collective works, for resale or redistribution to servers or
  lists, or reuse of any copyrighted component of this work in other works.
  }
\newcommand\copyrightnotice{%
\begin{tikzpicture}[remember picture,overlay]
\node[anchor=south,yshift=10pt] at (current page.south) {\fbox{\parbox{\dimexpr\textwidth-\fboxsep-\fboxrule\relax}{\copyrighttext}}};
\end{tikzpicture}%
}

\title{\LARGE \bf
Polygonal Obstacle Avoidance Combining \\ Model Predictive Control and Fuzzy Logic
}
\author{Michael Schröder$^{1}$, Eric Schöneberg$^{1}$, Daniel Görges$^{1}$ and Hans D. Schotten$^{2}$
\thanks{The authors gratefully acknowledge the financial support by the German Federal Ministry of Education and Research in the project Open6GHub (grant number: 16KISK004).}
\thanks{$^{1}$Institute of Electromobility, RPTU University Kaiserslautern-Landau, Kaiserslautern, Germany \{{\tt\small michael.schroeder, eric.schoeneberg, daniel.goerges}\}{\tt\small @rptu.de}}%
\thanks{$^{2}$Institute of Wireless Communications and Navigation, RPTU University Kaiserslautern-Landau, Kaiserslautern, Germany {\tt\small hans.schotten@rptu.de}}%
}

\begin{document}

\maketitle

\copyrightnotice

\thispagestyle{empty}
\pagestyle{empty}

\begin{abstract}
In practice, navigation of mobile robots in confined environments is often done using a spatially discrete cost-map to represent obstacles. Path following is a typical use case for model predictive control (MPC), but formulating constraints for obstacle avoidance is challenging in this case. Typically the cost and constraints of an MPC problem are defined as closed-form functions and typical solvers work best with continuously differentiable functions. This is contrary to spatially discrete occupancy grid maps, in which a grid's value defines the cost associated with occupancy. This paper presents a way to overcome this compatibility issue by re-formulating occupancy grid maps to continuously differentiable functions to be embedded into the MPC scheme as constraints. Each obstacle is defined as a polygon --  an intersection of half-spaces. Any half-space is a linear inequality representing one edge of a polygon. Using AND and OR operators, the combined set of all obstacles and therefore the obstacle avoidance constraints can be described. The key contribution of this paper is the use of fuzzy logic to re-formulate such constraints that include logical operators as inequality constraints which are compatible with standard MPC formulation. The resulting MPC-based trajectory planner is successfully tested in simulation. This concept is also applicable outside of navigation tasks to implement logical or verbal constraints in MPC.
\end{abstract}

\section{Introduction}\label{sec:Introduction}
For navigation of mobile robots, commonly the trajectory planner's input is a reference path, the current pose and a map of obstacles. A low-level controller then follows the planned trajectory. The Nav2 stack \cite{Nav2}, built on ROS2 \cite{ros2}, exemplifies this architecture. Using a Model Predictive Controller (MPC) for path tracking and trajectory planning while considering the kinematics of the robot is a common and suitable choice. MPC formulates the path tracking problem as a continuous optimization program, which is solved in a receding horizon fashion at run-time. The planned trajectory can deviate significantly from the reference path, especially if the reference path is generated without considering the robot's kinodynamic constraints. This necessitates constraints for obstacle avoidance in the MPC, despite a collision free reference path. In Nav2, an occupancy grid map represents the environment as a spatially discrete grid -- similar to a pixelated monochrome image -- where each cell indicates the presence of obstacles. To define the optimization problem, functions in closed form are needed instead. To transfer the pixelated map into this closed form, this paper proposes first extracting convex polygons from the map representing the obstacles. A set of convex polygons is universal and can account for obstacles that vary significantly in shape and size: e.g. humans, other robots or walls. The half-spaces defined by the edges of the polygons can then be combined using logical operators. These operators are realised using fuzzy logic to get continuously differentiable functions. The proposed MPC is implemented as a Nav2 controller plugin, with the open-source code \cite{iic_repo} shared to ensure reproducibility.
In the following, approaches of common Nav2 controllers regarding obstacle avoidance as well as MPC-based approaches for avoiding polygonal obstacles are reviewed. Furthermore it is discussed how fuzzy logic is currently used in MPC formulations. \\
The Nav2 controller DWB is based on the \textit{Dynamic Window Approach} (DWA) \cite{dwa_1997}, that considers discrete value combinations of linear and angular velocities leading to circular arcs as trajectory candidates. The trajectories that collide with an obstacle or lead to the robot not being able to stop in time before an obstacle are filtered out before choosing the velocity combination with the smallest cost.
The \textit{Regulated Pure Pursuit} controller \cite{macenski2023regulated}, the \textit{Graceful Motion Controller} (based on \cite{gracefulMotionController}) and the \textit{Vector Pursuit Controller} \cite{wit2000vector} move towards a look-ahead position (or pose), that is chosen from the reference path and is a predefined distance away. Some of them reduce their velocity while getting close to obstacles and check whether the trajectory is collision free before executing it, but they do not plan a trajectory around obstacles. They rely on staying close enough to the collision free reference path.
The \textit{Model Predictive Path Integral Controller} (MPPI) is described in \cite{MMPI} with the focus on the solver instead of the cost function (or constraints). That solver does not need a continuously differentiable cost function, because it calculates the cost for different possible trajectories (similar to DWA). A spatially discrete costmap is therefore no issue.
The \textit{Time Elastic Band} (TEB) controller \cite{ROSMANN2017142} is similar to a classical MPC formulation and suitable if the minimal distance to an obstacle is described by a continuous function. Despite the corresponding code depending on the repository \textit{costmap\_converter} \cite{Rosmann.2020}, this paper does not derive such a distance function for polygonal obstacles.
In conclusion, the current Nav2 controller do not rely on a closed form function to represent obstacles, since they do not use standard optimization solvers.
In \cite{zhang2020optimization} constraints for an optimization problem to avoid polygonal obstacles are proposed. The authors first define the distance (and later a \textit{signed distance}) between the robot and an obstacle as the result of a minimization problem. Using the dual problem of that, the distance is defined as the result of a maximization problem, that has to be above a predefined threshold. Doing so, it is sufficient to require that there exists a solution of the maximization problem that is above the threshold without actually solving the maximization problem. \cite{helling2021dual} builds upon that and modifies the constraints to improve the computation time. Both versions of this approach introduce Lagrange multipliers (for every edge of every obstacle of every time step) as additional decision variables. Similar to our approach, \cite{8796236} avoids obstacles where its edges are described by requiring that a function of the position is positive ($h_i(\matvar p)>0$). In contrast to our approach, combining these functions is done using $\prod_i \text{max}(h_i(\matvar p), 0)^2$ and requiring that this is zero for the robots position using a soft constraint. \cite{FMPC890326} is an example for the use of fuzzy logic for modeling the system for an MPC. \cite{mpcFuzzyDecisionFunctions} proposes a combination of MPC and fuzzy control, where all constraints and all goals (for all time steps) are expressed through membership functions. These are combined into one membership function that depends on the control inputs and is maximized to fulfill all goals and constraints as good as possible.

This paper proposes converting the spatially discrete cost-map first into polygons and then into a continuously differentiable function, so that obstacle constraints can be implemented while still using standard numerical solvers and without introducing additional optimization variables, which is its major contribution. More broadly, this can also be seen as an example use case of fuzzy logic in MPC constraints. With the proposed concept, any MPC formulation can be expanded by a fuzzy logic inspired constraint without adapting the obstacle avoidance motivation presented here.

The paper is organized as follows. First, \refSection{sec:MPC} describes the MPC problem for which the obstacle avoidance constraint is implemented. The half-spaces describing the polygonal obstacles are constructed according to \refSection{sec:halfspaces}. \refSection{sec:combining_halfspaces} and \refSection{sec:norming} are the main contribution of this paper, describing how all half-spaces of all polygons are combined to one function $g(\cdot) \in \mathcal{C}^\infty$. Implementing the corresponding constraint is described in \refSection{sec:Implementation}.
\section{Model Predictive Control}\label{sec:MPC}
The proposed trajectory planning is based on and tested with the Model Predictive Control (MPC) problem
\begin{subequations}\label{eq:MPCproblem.1}
    \begin{align}
        &\underset{\left\{\matvar{U}_k, \matvar{X}_k \right\}}{\text{minimize }}   && f_0 \left(\matvar{X}_k , \matvar{U}_k\right) \label{eq:MPCproblem.1a}\\
        &\text{subject to } && \matvar{x}_{k+i+1} = f\left(\matvar{x}_{k+i}, \matvar{u}_{k+i} \right)  \label{eq:MPCproblem.1b}\\
        &~                  && \matvar{x}_{k+i} \in \mathbb{X}_{k+i} \label{eq:MPCproblem.1c} \\
        &~                  && \matvar{u}_{k+i} \in \mathbb{U}_{k+i} \label{eq:MPCproblem.1d}
    \end{align}
\end{subequations}
with the state trajectory $\matvar{X}_k = \begin{pmatrix} \matvar{x}_{k+1}^\intercal & \matvar{x}_{k+2}^\intercal \dots & \matvar{x}_{k+N}^\intercal \end{pmatrix}^\intercal$ and the input trajectory $\matvar{U}_k = \begin{pmatrix} \matvar{u}_{k}^\intercal & \matvar{u}_{k+1}^\intercal \dots & \matvar{u}_{k+N-1}^\intercal \end{pmatrix}^\intercal$ where $\matvar{x}_{k+i} = \begin{pmatrix} \dots & \matvar{p}_{\text{bot},k+i}^\intercal & \dots \end{pmatrix}^\intercal$ is the state of the~robot including its position $\matvar p_\text{bot}\in\mathbb{R}^2$, $\matvar{u}_{k+i}$ is the input of the~robot, $k\in\mathbb{N}_0$ the current time, and $i\in [0, 1, \dotso , N-1]$ the time within the prediction horizon $N\in\mathbb{N}_{>0}$. The constraint \refEquation{eq:MPCproblem.1c} restricts $\matvar p_\text{bot}$ to avoid collisions with polygonally shaped obstacles, which is the focus of the paper. It furthermore limits the velocity and the constraint \refEquation{eq:MPCproblem.1d} primarily limits the acceleration. The cost function $f_0$ is mainly designed to let the robot follow a reference path. More details about the basic MPC formulation are described in \cite{schöneberg2025trajectoryplanningmodelpredictive}.
\section{Polygonal Obstacles}\label{sec:halfspaces}
Any obstacle $s$ is defined as a closed polygonal chain, which can be described by a set of convex hulls, e.g. utilizing the Graham Scan algorithm \cite{graham1972efficient}. Extracting the polygons from the costmap is not the focus of this paper. E.g.~in ROS2 this can be done with the package \textit{costmap\_converter} \cite{Rosmann.2020}. Any convex hull $\mathbb{S}_s$ with $V_s\in \mathbb{N}_{>2}$ vertices can be described by 
\begin{equation}\label{eq:convex hull}
    \mathbb{S}_s = \left\{\matvar{p} \in \mathbb{R}^2 \left| \matvar{b}_s - \matvar{A}_s \matvar{p} \succeq \matvar{0} \right. \right\}
\end{equation}
with $\matvar{A}_s \in \mathbb{R}^{V_s \times 2}$, $\matvar{b}_s \in \mathbb{R}^{V_s}$. The set of all obstacles is then $\mathbb{S} = \bigcup_s \mathbb{S}_s$. Every row of $\matvar{b}_s - \matvar{A}_s \matvar{p} \succeq \matvar{0}$ has to be understood as a separate inequality describing a half-space of $\mathbb{R}^2$, bounded by an edge of one of the polygons. All inequalities regarding one obstacle need to be fulfilled, so that $\matvar p \in \mathbb{S}$.
If the vertices of a polygon are given so that they move counterclockwise around the obstacle, the inequalities in (\ref{eq:convex hull}) can be constructed using
\begin{subequations}\label{eq:generate_matricex_from_verteces}
\begin{align}
    \matvar b_s &= \begin{pmatrix}
        x_{s,1} \Delta y_{s,1} - y_{s,1} \cdot \Delta x_{s,1}\\
        \vdots\\
        x_{s,V_s} \Delta y_{s,V_s} - y_{s,V_s} \cdot \Delta x_{s,V_s}\\
    \end{pmatrix}\\
    \matvar{A}_s &= \begin{bmatrix}
        \Delta y_{s,1} & -\Delta x_{s,1}\\
        \vdots & \vdots\\
        \Delta y_{s,V_s} & -\Delta x_{s,V_s}
    \end{bmatrix}\\
    \Delta x_{s,r} &= x_{s,(r+1)} - x_{s,r}\\ 
    \Delta y_{s,r} &= y_{s,(r+1)} - y_{s,r}\\ 
    \matvar{v}_{s,r} &=
    \begin{pmatrix}
        x_{s,r}\\
        y_{s,r}
    \end{pmatrix}
\end{align}
\end{subequations}
where $\matvar{v}_{s,r}$ is the vertex $r$ of the obstacle $s$. Despite $1 \leq r \leq V_s$, we define $\matvar v_{s,0} \coloneq \matvar v_{s,V_s}$ and $\matvar v_{s,V_s+1} \coloneq \matvar v_{s,1}$ (and similarly for $x_{s,r}$ and $y_{s,r}$), since the polygonal chain is closed.
%
\refEquation{eq:generate_matricex_from_verteces} is based on describing every edge of the polygon as
\begin{equation}
    f_{s,r}(x) = \frac{\Delta y_{s,r}}{\Delta x_{s,r}} (x - x_{s,r}) + y_{s,r}
\end{equation}

\begin{figure}[ht]
    \centering
    \begin{subfigure}[t]{0.49\linewidth}
        \centering
        \begin{tikzpicture}[scale=0.4]
        \def\fillEverythingBlue{\fill[fill=blue,opacity=0.3] (-20,-20) rectangle (20,20)}
        \clip (-4.8,-2.5) rectangle (1.7,4);
        \begin{scope}
            \clip (-10,-5) -- (10,5) -- (-10,5) -- cycle;
            \fillEverythingBlue{};
        \end{scope}
        \draw[draw=blue]  (-3,-1.5) -- (0,0);
        \draw[draw=red]   (0,0)     -- (-3,3);
        \draw[draw=green] (-3,3)    -- (-3,-1.5);
        \filldraw[black] (-3,-1.5) circle (2pt) node[left]{$v_{s,1}$};
        \filldraw[black] (0,0)     circle (2pt) node[anchor=west]{$v_{s,2}$};
        \filldraw[black] (-3,3)    circle (2pt) node[left]{$v_{s,3}$};
        \end{tikzpicture}
        \caption{Halfspace defined by edge below polygon}
        \label{fig:halfspace_inequalities_below}
    \end{subfigure}
    \begin{subfigure}[t]{0.49\linewidth}
        \centering
        \begin{tikzpicture}[scale=0.4]
        \def\fillEverythingRed{\fill[fill=red,opacity=0.3] (-20,-20) rectangle (20,20)}
        \clip (-4.8,-2.5) rectangle (1.7,4);
        \begin{scope}
            \clip (-10,10) -- (10,-10) -- (-10,-10) -- cycle;
            \fillEverythingRed{};
        \end{scope}
        \draw[draw=blue]  (-3,-1.5) -- (0,0);
        \draw[draw=red]   (0,0)     -- (-3,3);
        \draw[draw=green] (-3,3)    -- (-3,-1.5);
        \filldraw[black] (-3,-1.5) circle (2pt) node[left]{$v_{s,1}$};
        \filldraw[black] (0,0)     circle (2pt) node[anchor=west]{$v_{s,2}$};
        \filldraw[black] (-3,3)    circle (2pt) node[left]{$v_{s,3}$};
        \end{tikzpicture}
        \caption{Halfspace defined by edge above polygon}
        \label{fig:halfspace_inequalities_above}
    \end{subfigure}
    \caption{Halfspace inequalities}
    \label{fig:halfspace_inequalities}
\end{figure}
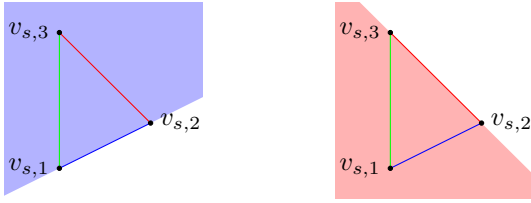

For the edges below the polygon (see \refFigure{fig:halfspace_inequalities_below}), $\Delta x_{s,r}>0$ and the corresponding half-spaces are described by $p_y \geq f_{s,r}(p_x)$. The half-spaces defining the edges above the polygon (see \refFigure{fig:halfspace_inequalities_above}) are described by $p_y \leq f_{s,r}(p_x)$ and in that case $\Delta x_{s,r}<0$. Both cases can be described by
\begin{align}
    & p_y\cdot \Delta x_{s,r} \geq f_{s,r}(p_x)\cdot \Delta x_{s,r}\\
    &p_y\cdot \Delta x_{s,r} - \Delta y_{s,r}(p_x - x_{s,r}) - y_{s,r} \Delta x_{s,r} \geq 0
\end{align}
leading to \refEquation{eq:generate_matricex_from_verteces}, which is also valid for cases with $\Delta x_{s,r} = 0$. Alternatively this can be derived similarly to \cite{helling2021dual} by stating
\begin{align}
    \begin{pmatrix}
        \Delta y_{s,r}\\
        -\Delta x_{s,r}
    \end{pmatrix}
    \cdot (\matvar p - \matvar v_r)^\intercal < 0
\end{align}
where the first vector is orthogonal to the edge and points to the outside of the polygon. This also leads to \refEquation{eq:generate_matricex_from_verteces} and has the advantage of also being applicable for describing planes of three-dimensional obstacles.

\section{Combining Halfspaces using Fuzzy Logic} \label{sec:combining_halfspaces}
Any row $r$ of \refEquation{eq:convex hull} and \refEquation{eq:generate_matricex_from_verteces} is a separate condition connected with a logical AND. The intersection of all the halfspaces of polygon $s$ is
\begin{equation}
    \mathbb{S}_s = \left\{ \matvar p \in \mathbb{R}^2 \left| 
    \bigwedge_r (b_{s,r} - \matvar{a}_{s,r} \matvar{p} \geq 0)
    \right. \right\} \label{eq:merging_halfspaces}
\end{equation}
where $\matvar{a}_{s,r}$ and $b_{s,r}$ are the rows $r$ of $\matvar{A}_s$ and $\matvar{b}_s$. The expression $\mathbb{S} = \bigcup_s \mathbb{S}_s$ for merging all the polygons can be stated as
\begin{subequations}
\begin{align}
    \mathbb{S} &= \left\{ \matvar p \in \mathbb{R}^2 \left|
    \bigvee_s \left( \matvar p \in \mathbb{S}_s \right)
    \right. \right\} \label{eq:merging_polygons}\\
    &= \left\{ \matvar p \in \mathbb{R}^2 \left|
    \bigvee_s \left( \bigwedge_r (b_{s,r} - \matvar{a}_{s,r} \matvar{p} \geq 0) \right)
    \right. \right\} \label{eq:merging_halfspaces_and_polygons_logically}
\end{align}
\end{subequations}
$\matvar p_{\text{bot}} \notin \mathbb{S}$, i.e.~the robot not colliding with an obstacle, can not be easily implemented using \refEquation{eq:merging_halfspaces_and_polygons_logically} as a constraint in an MPC problem that should be formulated using a standard solvers like e.g.~IPOPT \cite{ipopt}. A closed-form expression of the logical AND and OR using continuous and differentiable functions is desirable since ideally a numerical solver for continuous optimization problems should be used for computational efficiency. Fuzzy logic seems appropriate for this purpose. This starts with the fuzzyfication of the statements $b_{s,r} - \matvar{a}_{s,r} \matvar{p} \geq 0$. The sigmoid function
\begin{equation}
    \signum(x) = \frac{1}{1+e^{-x}}
\end{equation}
is used as a membership function since $\signum \in \mathcal{C}^\infty$. After adding a scaling factor $c>0$, $\signum(c(b_{s,r} - \matvar{a}_{s,r}\matvar{p}))$ represents whether point $\matvar p$ is inside the corresponding half-space. Next, the logic operations in \refEquation{eq:merging_halfspaces_and_polygons_logically} are realised. In fuzzy logic, multiplication can be used as an AND operator. For this application, representing an OR operation with an addition is sufficient. Utilizing all that leads to the following approximation of \refEquation{eq:merging_halfspaces_and_polygons_logically}:
\begin{align}
    g(\matvar p) &= \sum_s \prod_v \signum(c(b_{s,r} - \matvar{a}_{s,r} \matvar{p}))\\
    \mathbb{S} &\approx \left\{\matvar{p} \in \mathbb{R}^2 \left| g(\matvar p) > 0.5 \right. \right\} \label{eq:obstacle_set_approximation}
\end{align}
To avoid collisions with the obstacles, one demands $\matvar p_{\text{bot}} \notin \mathbb{S}$, leading to the inequality constraint
\begin{equation}\label{eq:static_obst_constraint}
    g(\matvar{p}_{\text{bot},k+i}) < 0.5 \;\; \forall 1\leq i \leq N
\end{equation}
There are also other fuzzy operators to choose from. E.g. $b \lor c$ can also be represented by $\text{max}(b, c)$, but $\text{max}(\cdot) \notin \mathcal{C}^1$ and is therefore inappropriate for the use with gradient-based solvers. Commutative and associative operators are preferred since the order in which the detected polygons are listed can change between the MPC cycles, which should not have an influence on the set of feasible positions. An OR operation $a = b \vee c$ is often realized with $\mu(a) = \mu(b) + \mu(c) -(\mu(b) \cdot \mu(c))$, where $\mu$ is the fuzzified representation of a statement. The term $ -(\mu(b) \cdot \mu(c))$ ensures the property $0<\mu(a)<1$, which is not needed for this application. Furthermore this term increases the complexity and therefore the computation time, especially if four or more statements are concatenated with an OR. Hence, the alternative definition $\mu(a) = \mu(b) + \mu(c)$ is chosen, which in the end results in slightly more conservative constraints. Negating a statement is typically realised with $\mu(\overline{a})=1-\mu(a)$, which can be used to explain \refEquation{eq:static_obst_constraint}.

The threshold of $0.5$ in \refEquation{eq:obstacle_set_approximation} and \refEquation{eq:static_obst_constraint} seems intuitive considering that $\signum(0)=0.5$ and $0$ and $1$ represent FALSE and TRUE. \refFigure{fig:static_obst_without_norming_allowed_area} shows, that the constraint \refEquation{eq:static_obst_constraint} is not conservative enough at the vertices. At the vertices, two edges of the polygon intersect and $g(\matvar v_{s,r}) \approx \signum(0) \cdot \signum(0) = 0.25$ if the influence of the other edges is neglected. To better include the vertices in $\mathbb{S}_s$, the threshold in \refEquation{eq:static_obst_constraint} is changed from $0.5$ to $0.25$. With a threshold of $0.25$ the allowed area behaves as desired at the vertices of the red polygon in \refFigure{fig:static_obst_threshold_025_allowed_area} and is a bit conservative at the edges in between. An even smaller threshold makes the constraints more conservative. It is justified if soft constraints are used and a buffer for the slag variable is desired or if more than two sigmoid terms (each corresponding to one edge) influence $g(\matvar{p})$ significantly at some point of a polygon.
For three-dimensional obstacles like polyhedrons -- where three planes intersect in vertices -- a threshold of $0.5^3$ would be chosen.
\section{Normalization of the Conditions} \label{sec:norming}
Each row within $\matvar{A}_s \matvar{p} - \matvar{b}_s \succeq \matvar{0}$ can be scaled by an arbitrary positive factor without changing the halfspace that it describes. Therefore how steep the transition from 0 to 1 is can vary significantly at different edges. This undesirable effect is shown in \refFigure{fig:static_obst_without_norming}, where the flat slope of the left edge of the right polygon influences other edges. For demonstration purposes, the corresponding row was scaled accordingly. For a position $\matvar p_{\text{edge}}$ that is close to edge $r$ of obstacle $s$ and far away from all the other edges, we can approximate
\begin{equation}
    g(\matvar p_{\text{edge}}) \approx  \signum(c(b_{s,r} - \matvar{a_{s,r}} \cdot \matvar{p_{\text{edge}}}))
\end{equation}
The slope of the transition at the edge depends on
\begin{equation}
    \left.\frac{dg(\matvar p)}{d \matvar{p}}\right\rvert_{\matvar p=\matvar{p_{\text{edge}}}} = \underbrace{\left.\frac{d\, \signum(x)}{dx}\right\rvert_{x=0}}_{const} \cdot (- \matvar{a_{s,r}}c)
\end{equation}
To guarantee a similar slope at all edges, every row $r$ is normalized
\begin{equation}
    \matvar{a}_{s,r}' = \frac{\matvar{a}_{s,r}}{\lVert \matvar{a}_{s,r} \rVert_2}
    \hspace{1cm}
    b_{s,r}' = \frac{b_{s,r}}{\lVert \matvar{a}_{s,r} \rVert_2}
\end{equation}
The desired slope is set using the design parameter $c$. This leads to a more consistent and predictable form of $g(\matvar p)$ and the absolute value of the resulting expression
\begin{align}
    |b_{s,r}' - \matvar{a}_{s,r}' \cdot \matvar{p}|
    = \frac{\left\lvert \begin{pmatrix} \Delta x_{s,r}\\\Delta y_{s,r}\\0 \end{pmatrix} \times \begin{pmatrix} (\matvar p - \matvar{v}_{s,r})\\0\end{pmatrix}  \right\rvert}{\lVert \matvar{a}_{s,r} \rVert_2}
\end{align}
can be interpreted as the distance from $\matvar p$ to the polygon's edge (edge from $\matvar v_{s,r}$ to $\matvar v_{s,r+1}$) as shown in \cite{distanceLineAndPoint}.
\ifthenelse{\boolean{showFigures}}{
    \begin{figure}[ht]                  
        \centering
        \begin{subfigure}[b]{0.49\columnwidth}
            \centering
            \ifthenelse
            {\boolean{use_tikz_instead_of_pdf}}
            {\input{Figures/3D_2D_comparison/3D_plot_without_norming_flat_slope.tex}}
            {\includegraphics[width=\textwidth]{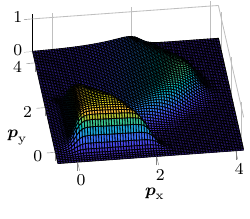}}
            \caption{$g(\matvar p)$}
            \label{fig:static_obst_without_norming_3d}
        \end{subfigure}
        \begin{subfigure}[b]{0.49\columnwidth}
            \centering
            \includegraphics[width=\textwidth]{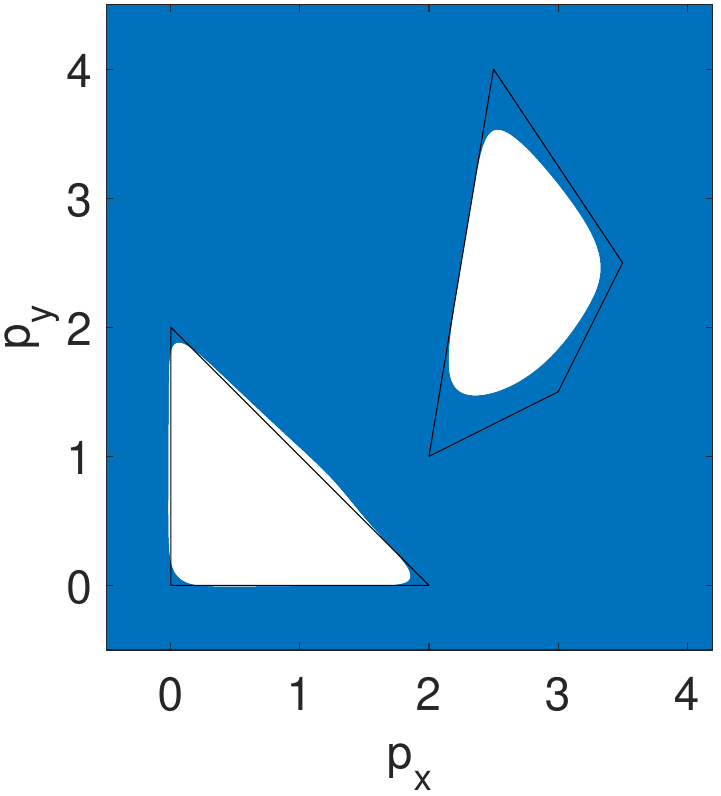}
            \caption{Allowed area and desired polygons}
            \label{fig:static_obst_without_norming_allowed_area}
        \end{subfigure}
        \caption{Example of static obstacles without normalization\\(with $c=6$ and threshold of 0.5 in \refEquation{eq:static_obst_constraint})}
        \label{fig:static_obst_without_norming}
    \end{figure}
}{}

\section{Polygon Preprocessing}

\ifthenelse{\boolean{showFigures}}{
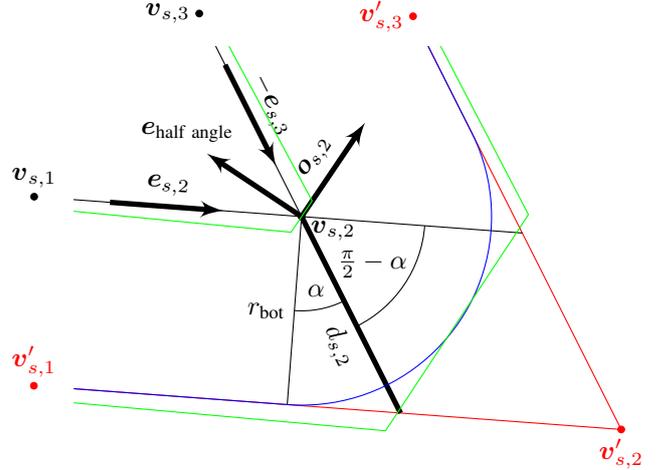
\begin{figure}[ht]
    \centering
    \begin{tikzpicture}[scale=2.5]
    \def\scalingArrow{0.6}
    \def\ax{1.0}
    \def\ay{2.3}
    \def\bx{5.0}
    \def\by{2.0}
    \def\cx{3.5}
    \def\cy{5.0}
    \def\r{1.0};
    %
    \def\eaabs{sqrt((\ax-\cx)^2+(\ay-\cy)^2)};
    \def\eax{(\ax-\cx)/\eaabs};
    \def\eay{(\ay-\cy)/\eaabs};
    \def\ebabs{sqrt((\bx-\ax)^2+(\by-\ay)^2)};
    \def\ebx{(\bx-\ax)/\ebabs};
    \def\eby{(\by-\ay)/\ebabs};
    \def\ecabs{sqrt((\cx-\bx)^2+(\cy-\by)^2)};
    \def\ecx{(\cx-\bx)/\ecabs};
    \def\ecy{(\cy-\by)/\ecabs};
    %
    \def\alphaa{90-acos(-\eax*\ebx-\eay*\eby)};
    \def\da{\r/cos(\alphaa)};
    \def\Ax{\ax+\da*(\eax-\ebx)};
    \def\Ay{\ay+\da*(\eay-\eby)};
    \def\alphab{90-acos(-\ebx*\ecx-\eby*\ecy)};
    \def\db{\r/cos(\alphab)};
    \def\Bx{\bx+\db*(\ebx-\ecx)};
    \def\By{\by+\db*(\eby-\ecy)};
    \def\alphac{90-acos(-\ecx*\eax-\ecy*\eay)};
    \def\dc{\r/cos(\alphac)};
    \def\Cx{\cx+\dc*(\ecx-\eax)};
    \def\Cy{\cy+\dc*(\ecy-\eay)};
    \begin{scope}
        \clip (3.8,0.6) rectangle (6.9,2.9);
        \draw (\ax,\ay) -- (\bx,\by) -- (\cx,\cy) -- (\ax,\ay);
        \filldraw (\ax,\ay) circle (0.5pt) node[left] {$\matvar{v}_{s,1}$};
        \filldraw (\bx,\by) circle (0.5pt) node[below right,yshift=2pt] {$\matvar{v}_{s,2}$};
        \filldraw (\cx,\cy) circle (0.5pt) node[above] {$\matvar{v}_{s,3}$};
        %
        \def\eaShift{1.5};
        \draw[-latex', line width=2pt] ({\cx+\eaShift*\eax},{\cy+\eaShift*\eay}) -- ({\cx+\eax*\scalingArrow+\eaShift*\eax},{\cy+\eay*\scalingArrow+\eaShift*\eay}) node[midway, above, sloped] {$\matvar{e}_{s,1}$};
        \def\ebShift{3.0};
        \draw[-latex', line width=2pt] ({\ax+\ebShift*\ebx},{\ay+\ebShift*\eby}) -- ({\ax+\ebx*\scalingArrow+\ebShift*\ebx},{\ay+\eby*\scalingArrow+\ebShift*\eby}) node[midway, above, sloped] {$\matvar{e}_{s,2}$};
        \def\ecShift{0.3};
        \draw[latex'-, line width=2pt] ({\bx+\ecShift*\ecx},{\by+\ecShift*\ecy}) -- ({\bx+\ecx*\scalingArrow+\ecShift*\ecx},{\by+\ecy*\scalingArrow+\ecShift*\ecy}) node[midway, above, sloped] {$-\matvar{e}_{s,3}$};
        %
        \draw[red] ({\Ax},{\Ay}) -- ({\Bx},{\By}) -- ({\Cx},{\Cy}) -- ({\Ax},{\Ay});
        \filldraw[red] ({\Ax},{\Ay}) circle (0.5pt) node[below] {$\matvar{v}_{s,1}'$};
        \filldraw[red] ({\Bx},{\By}) circle (0.5pt) node[below] {$\matvar{v}_{s,2}'$}; 
        \filldraw[red] ({\Cx},{\Cy}) circle (0.5pt) node[above] {$\matvar{v}_{s,3}'$};
        %
        \draw ({\bx},{\by}) -- ({\bx+\db*\ebx},{\by+\db*\eby}) node[at end](intersectionOnRight){};
        \draw[line width=2pt] ({\bx},{\by}) -- ({\bx-\db*\ecx},{\by-\db*\ecy}) node[at end](intersectionD){} node[midway, below, xshift=0.3cm, sloped] {$d_{s,2}$};
        \draw ({\bx},{\by}) -- ($({\Ax},{\Ay})!({\bx},{\by})!({\Bx},{\By})$) node[at end](intersectionR){} node[midway, left] {$r_{\text{bot}}$};
        \coordinate (b) at ({\bx},{\by});
        \tkzMarkAngle[mark=none,size=0.5cm](intersectionR,b,intersectionD)
        \tkzLabelAngle[pos=0.4](intersectionR,b,intersectionD){$\alpha$}
        \tkzMarkAngle[mark=none,size=0.65cm](intersectionD,b,intersectionOnRight)
        \tkzLabelAngle[pos=0.45](intersectionD,b,intersectionOnRight){$\frac{\pi}{2}-\alpha$}
        %
        \coordinate (A) at ({\Ax},{\Ay});
        \coordinate (B) at ({\Bx},{\By});
        \coordinate (C) at ({\Cx},{\Cy});
        \coordinate (a) at ({\ax},{\ay});
        \coordinate (b) at ({\bx},{\by});
        \coordinate (c) at ({\cx},{\cy});
        \def\distanceFromVTwo{sqrt( pow(veclen(\Bx-\bx,\By-\by),2) - pow(\r,2) )}
        \begin{scope}
            \clip (B) circle ({\distanceFromVTwo});
            \draw[blue] (b) circle (\r);
        \end{scope}
        \draw[blue] (A) -- ({\Bx-\distanceFromVTwo*\ebx},{\By-\distanceFromVTwo*\eby});
        \draw[blue] (C) -- ({\Bx+\distanceFromVTwo*\ecx},{\By+\distanceFromVTwo*\ecy});
        %
        \def\minDistanceBetweenPoints{0.2}
        \def\winkelhalbierendeX{(-\ebx+\ecx) / veclen(-\ebx+\ecx, -\eby+\ecy)};
        \def\winkelhalbierendeY{(-\eby+\ecy) / veclen(-\ebx+\ecx, -\eby+\ecy)};
        \def\orthogonalVecX{\winkelhalbierendeY};   
        \def\orthogonalVecY{-\winkelhalbierendeX};
        \draw[-latex', line width=2pt] ({\bx},{\by}) -- ({\bx+\scalingArrow*\winkelhalbierendeX},{\by+\scalingArrow*\winkelhalbierendeY}) node[above left, xshift=0.5cm] {$\matvar e_{\text{half angle}}$};
        \draw[-latex', line width=2pt] ({\bx},{\by}) -- ({\bx+\scalingArrow*\orthogonalVecX},{\by+\scalingArrow*\orthogonalVecY}) node[midway, above, sloped] {$\matvar{o}_{s,2}$};
        \def\bNewOneX{(\bx-0.5*\minDistanceBetweenPoints*\orthogonalVecX)}; 
        \def\bNewOneY{(\by-0.5*\minDistanceBetweenPoints*\orthogonalVecY)};
        \def\bNewTwoX{(\bx+0.5*\minDistanceBetweenPoints*\orthogonalVecX)};
        \def\bNewTwoY{(\by+0.5*\minDistanceBetweenPoints*\orthogonalVecY)};
        \draw[green] (a) -- ({\bNewOneX},{\bNewOneY}) -- ({\bNewTwoX},{\bNewTwoY}) -- (c);
        \def\ebNewOneX{(\bNewOneX-\ax) / veclen(\bNewOneX-\ax,\bNewOneY-\ay)};
        \def\ebNewOneY{(\bNewOneY-\ay) / veclen(\bNewOneX-\ax,\bNewOneY-\ay)};
        \def\ebNewTwoX{(\bNewTwoX-\bNewOneX) / veclen(\bNewTwoX-\bNewOneX,\bNewTwoY-\bNewOneY)};
        \def\ebNewTwoY{(\bNewTwoY-\bNewOneY) / veclen(\bNewTwoX-\bNewOneX,\bNewTwoY-\bNewOneY)};
        \def\ecNewX{(\cx-\bNewTwoX) / veclen(\cx-\bNewTwoX,\cy-\bNewTwoY)};
        \def\ecNewY{(\cy-\bNewTwoY) / veclen(\cx-\bNewTwoX,\cy-\bNewTwoY)};
        \def\alphabNewOne{90-acos(-\ebNewOneX*\ebNewTwoX-\ebNewOneY*\ebNewTwoY)};
        \def\dbNewOne{\r/cos(\alphabNewOne)};
        \def\BNewOneX{\bNewOneX+\dbNewOne*(\ebNewOneX-\ebNewTwoX)};
        \def\BNewOneY{\bNewOneY+\dbNewOne*(\ebNewOneY-\ebNewTwoY)};
        \def\alphabNewTwo{90-acos(-\ebNewTwoX*\ecNewX-\ebNewTwoY*\ecNewY)};
        \def\dbNewTwo{\r/cos(\alphabNewTwo)};
        \def\BNewTwoX{\bNewTwoX+\dbNewTwo*(\ebNewTwoX-\ecx)};
        \def\BNewTwoY{\bNewTwoY+\dbNewTwo*(\ebNewTwoY-\ecy)};
        \draw[green] (A) -- ({\BNewOneX},{\BNewOneY}) -- ({\BNewTwoX},{\BNewTwoY}) -- (C);
    \end{scope}
    %
    \def\shownVOneShift{2.6};
    \def\shownVOneX{\ax+\shownVOneShift*\ebx};
    \def\shownVOneY{\ay+\shownVOneShift*\eby};
    \filldraw ({\shownVOneX}, {\shownVOneY}) circle (0.5pt) node[above] {$\matvar{v}_{s,1}$};
    \def\shownVOneDashShift{-3.1};
    \def\shownVOneDashX{\Bx+\shownVOneDashShift*\ebx};
    \def\shownVOneDashY{\By+\shownVOneDashShift*\eby};
    \filldraw[red] ({\shownVOneDashX}, {\shownVOneDashY}) circle (0.5pt) node[above] {$\matvar{v}_{s,1}'$};
    \def\shownVThreeShift{2.15};
    \def\shownVThreeX{\cx-\shownVThreeShift*\ecx};
    \def\shownVThreeY{\cy-\shownVThreeShift*\ecy};
    \filldraw ({\shownVThreeX}, {\shownVThreeY}) circle (0.5pt) node[left] {$\matvar{v}_{s,3}$};
    \def\shownVThreeDashShift{-2.45};
    \def\shownVThreeDashX{\Bx-\shownVThreeDashShift*\ecx};
    \def\shownVThreeDashY{\By-\shownVThreeDashShift*\ecy};
    \filldraw[red] ({\shownVThreeDashX}, {\shownVThreeDashY}) circle (0.5pt) node[left] {$\matvar{v}_{s,3}'$};
\end{tikzpicture}
    \caption{Polygon preprocessing}
    \label{fig:polygon_preprocessing}
\end{figure}
}{}
%
The robot is conservatively approximated by a circle with an radius $r_{\text{bot}}$. Collisions with obstacles are prevented if the centre of the robot stays outside of the area that is marked blue in \refFigure{fig:polygon_preprocessing}. This can be achieved by shifting the edges by $r_{\text{bot}}$. The corresponding vertices $\matvar{v}_{s,r}'$ of the {\color{red} inflated polygon} are calculated using
\begin{subequations}
\begin{align}
    \matvar{e}_{s,r} &\coloneq \frac{\matvar{v}_{s,r} - \matvar{v}_{s,r-1}}{\left\lVert \matvar{v}_{s,r} - \matvar{v}_{s,r-1} \right\rVert_2}\\ 
    \cos\left(\frac{\pi}{2}-\alpha_{s,r}\right) &= \matvar{e}_{s,r}^\intercal \cdot (-\matvar{e}_{s,r+1})\\
    d_{s,r} &= \frac{r_{\text{bot}}}{\cos (\alpha_{s,r})} \label{eq:shifting_vertecies_d}\\
    \matvar{v}_{s,r}' &= \matvar{v}_{s,r} + d_{s,r}\matvar{e}_{s,r} - d_{s,r}\matvar{e}_{s,r+1}
\end{align} \label{eq:shifting_vertecies}
\end{subequations}
The results so far can be seen in \refFigure{fig:static_obst_threshold_025}.
\ifthenelse{\boolean{showFigures}}{
    \begin{figure}[ht]
        \centering
        \begin{subfigure}[b]{0.49\columnwidth}
            \centering
            \ifthenelse
            {\boolean{use_tikz_instead_of_pdf}}
            {\input{Figures/3D_2D_comparison/3D_plot_threshold_025}}
            {\includegraphics[width=\textwidth]{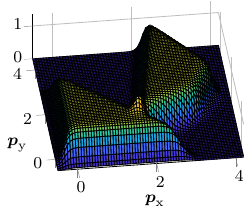}}
            \caption{$g(\matvar p)$}
            \label{fig:static_obst_threshold_025_allowed_area_3d}
        \end{subfigure}
        \begin{subfigure}[b]{0.49\columnwidth}
            \centering
            \includegraphics[width=\textwidth]{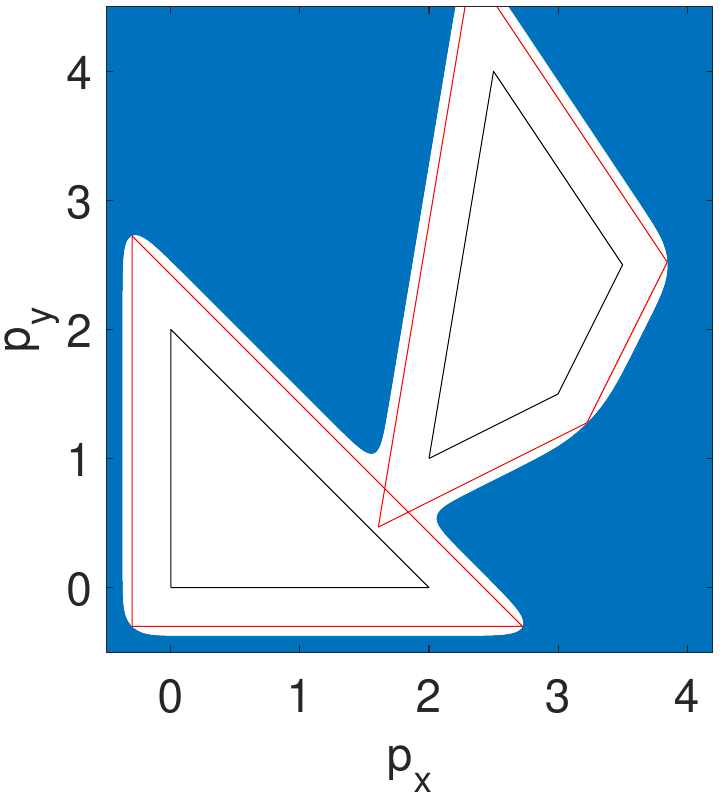}
            \caption{Allowed area and desired polygons}
            \label{fig:static_obst_threshold_025_allowed_area}
        \end{subfigure}
        \caption{inflated polygons (with threshold in \refEquation{eq:static_obst_constraint} of $0.25$, normalization, $c=15$ and $r_{\text{bot}}=0.3$)}
        \label{fig:static_obst_threshold_025}
    \end{figure}
}{}
If $\frac{\pi}{2}-\alpha$ is very small, the point $\matvar v_{s,r}'$ is moved  far away from $\matvar v_{s,r}$ leading to unnecessarily restrictive constraints. This can be prevented by replacing the corresponding $\matvar v_{s,r}$ with the two points
\begin{subequations}
\begin{align}
    \matvar v_{s,r,\text{new}} &= \matvar v_{s,r} \mp 0.5 \cdot d_{\text{min}} \cdot \matvar o_{s,r}\\
    \matvar o_{s,r} &= \begin{pmatrix} e_{\text{half angle},y}\\ -e_{\text{half angle},x} \end{pmatrix}\\
    \matvar e_{\text{half angle}} &= \frac{-\matvar e_{s,2}+\matvar e_{s,3}}{|-\matvar e_{s,2}+\matvar e_{s,3}|}
\end{align}
\end{subequations}
where $d_{\text{min}}$ is the minimum distance between the vertices of the polygons. $\matvar o_{s,r}$ is $\matvar e_{\text{half angle}}$ rotated clockwise by~$90^\circ$. Afterwards, \refEquation{eq:shifting_vertecies} is applied leading to the outer green polygon in \refFigure{fig:polygon_preprocessing} which is less restrictive in the corner in comparison to the red polygon.


%

\section{Implementation}\label{sec:Implementation}

While  defining the optimization problem in CasADi \cite{Andersson2018}, one has to already know the number of obstacles ($O$) and the number of edges of every obstacle ($V_s$). Since these numbers change during runtime, upper limits of them are assumed ($\hat{O}$ and $\hat{V}$) and \refEquation{eq:obstacle_set_approximation} is modified to
\begin{align}
    g(\matvar p) &= \sum_{s=1}^{\hat{O}} \left[ c_s \cdot \prod_{r=1}^{\hat{V}} \left( (1-d_{s,r}) + d_{s,r} \cdot g_{s,r}(\matvar p) \right) \right]\\
    g_{s,r}(\matvar p) &= \signum (c(b_{s,r}' - \matvar{a}_{s,r}' \cdot \matvar{p}))
\end{align}
If there are less obstacles ($O < \hat{O}$), the unused terms of $g(\cdot)$ are deactivated by setting the parameters \mbox{$c_1 \ldots c_O = 1$} and $c_{O+1} \ldots c_{\hat{O}} = 0$. If $O \geq \hat{O}$, all $c_{1} \ldots c_{\hat{O}} = 1$ and only the closest obstacles are considered in the constraints. Similarly we set $d_{s,1} \ldots d_{s,V_s} = 1$ and $d_{s,V_s+1} \ldots d_{s,\hat{V}} = 0$, if obstacle $s$ has less than $\hat{V}$ edges. These parameters can be set during every MPC cycle without changing the problem definition. The constraint \refEquation{eq:static_obst_constraint} is implemented as soft constraints using
\begin{align}
    g(\matvar{p}_{k+i}) - \epsilon_{\text{slag}} < 0.25 \qquad \forall{1<i<N}
\end{align}

\section{Evaluation}
\begin{figure}[ht]
    \centering
    \includegraphics[trim={1.0cm 0.0cm 0.0cm 0.5cm},clip,width=0.8\linewidth] {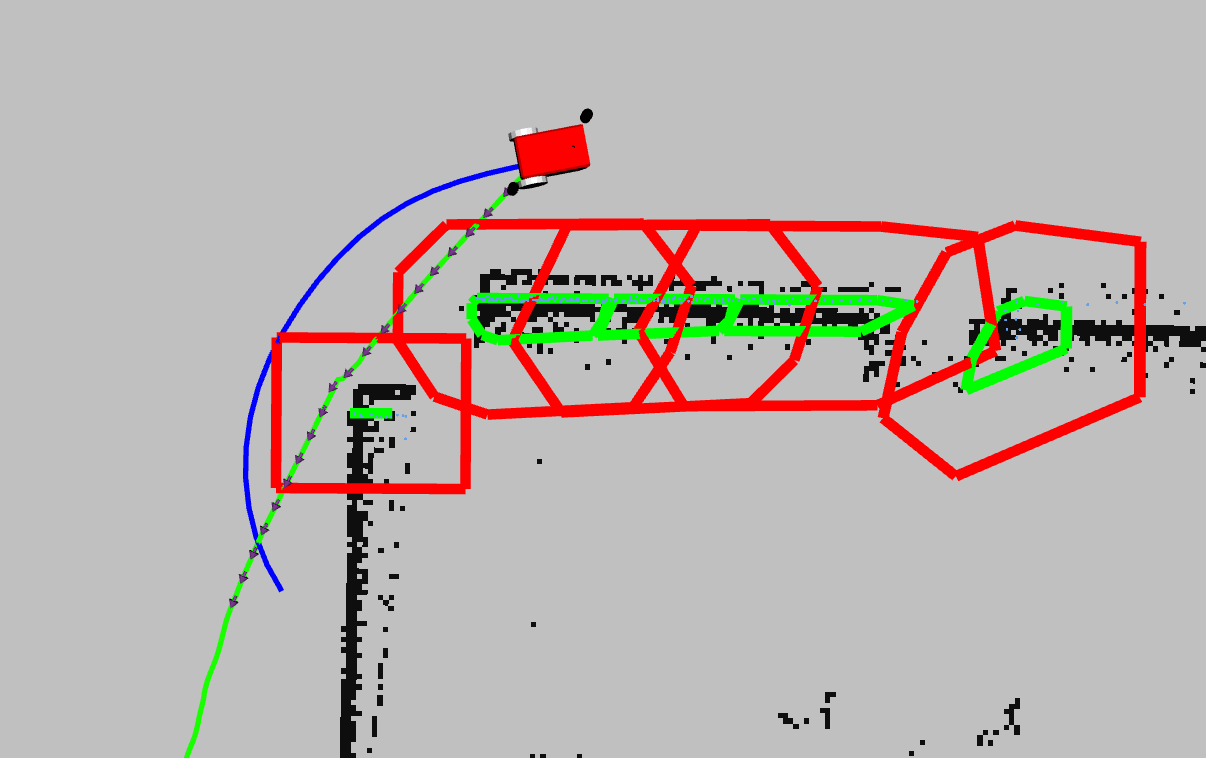}
    \caption{Deviation from the reference path to avoiding obstacles}
    \label{fig:evaluation}
\end{figure}

The proposed controller was tested in the simulation environment Gazebo \cite{gazebo}. To better demonstrate the effect of the proposed constraints, in \refFigure{fig:evaluation} the path planner was intentionally misconfigured so that it generated a reference path that is too close to obstacles. The controller successfully plans a trajectory (blue) around the inflated obstacles (red).

\begin{figure}[ht]
    \centering
    \begin{subfigure}[b]{0.49\columnwidth}
        \centering
        \includegraphics[trim={2.0cm 1.4cm 0.5cm 0.55cm},clip,height=3.35cm] {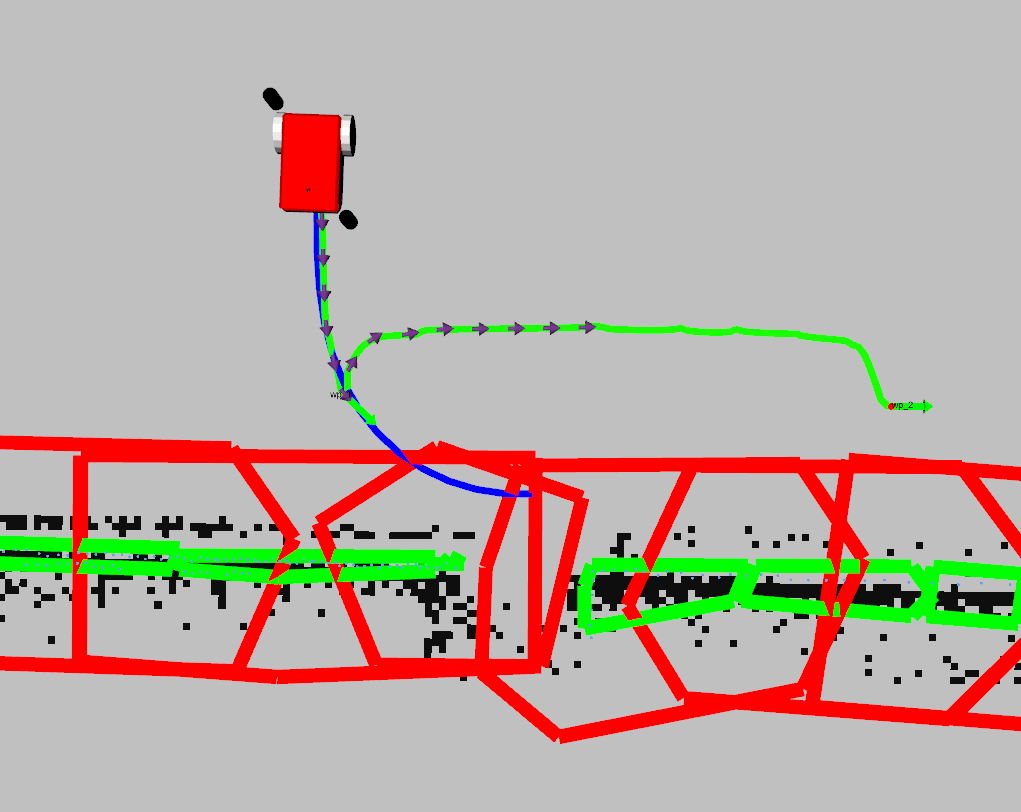} 
        \caption{without constraints ($\hat{O}=0$)}
        \label{fig:sim_corner_without_constraints}
    \end{subfigure}
    \begin{subfigure}[b]{0.49\columnwidth}
        \centering
        \includegraphics[trim={0.0cm 0.0cm 0.0cm 0.0cm},clip,height=3.35cm] {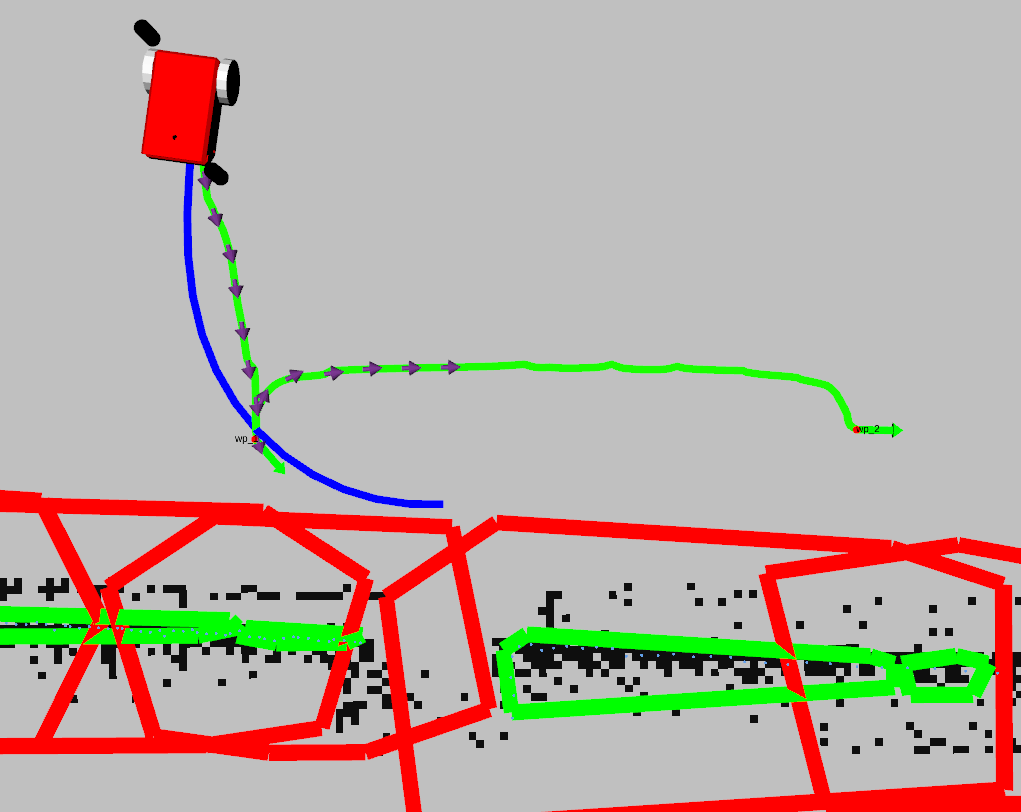}
        \caption{with constraints ($\hat{O}=10$)}
        \label{fig:sim_corner_with_constraints}
    \end{subfigure}
    \caption{Avoid collisions while driving around corners}
    \label{fig:sim_corner}
\end{figure}

\refFigure{fig:sim_corner} demonstrates the necessity of the proposed constraints. In \refFigure{fig:sim_corner_without_constraints} the constraints were deactivated for comparison.

Currently, the polygons are generated every cycle and therefore change slightly in between the cycles. At every time step, the polygons are reasonable, but changing the feasible set is undesirable. An initial guess that is based on the results of the previous cycle may be infeasible. The implementation as soft constraints and therefore adding a safety distance to $r_{\text{bot}}$ in \refEquation{eq:shifting_vertecies_d} helped mitigating this problem. A more consistent generation of the polygons could decrease the computation time and lead to a smoother motion of the robot.

The simulations are executed on an Intel i7-10700K. In the worst case scenario, where the robot drove next to obstacles with a misconfigured path planner (like in \refFigure{fig:evaluation}), the computation time was between $10ms$ and $25ms$ for most cycles. With appropriate settings of the path planner, the computation time was between $10ms$ and $16ms$ with occasional outliers. The simulations were performed with $\hat{O}=3$, $\hat{V}=8$ and a scaling factor $c=7.0$.
If the scaling factor is chosen too small (e.g. $c=1.0$), the robot drives through the polygons. How such a flat slope of $g(\matvar p)$ makes the constraints less conservative, is discussed in the context of \refFigure{fig:static_obst_without_norming}. If the transition between 0 and 1 is too steep, the robot starts driving through the polygons too (e.g. with $c=90$) or error messages from the solver occur (e.g. with $c=150$) indicating numerical problems. It is plausible, that in such situations, the initial guess includes a position, that violates the constraint ($g(\matvar p) \approx 1$) and the gradient of $g(\matvar p)$ is very small if $c$ is too large and therefore the transition width is too narrow. Together with the soft constraints this could lead to local minima. For getting a first working value of $c$, one should plot $\signum(cx)$ and select $c$ so that the width of the transition seems appropriate. Initial guesses that violate $g(\matvar p) < 0.25$ too much (outside of the transition width) should be avoided.
\section{Conclusion and Future Work}
The proposed controller has proven to work in a simulation environment that is very close to the real world. Improvements regarding the computation time and comparison with Nav2 controllers and the approaches of \cite{zhang2020optimization} and \cite{8796236} are the next steps in future work.
%
This paper introduces a concept that can be interpreted from two different perspectives. It is a systematic curve fitting to get a conservative and smooth closed-form approximation of the discrete costmap to enable the formulation of constraints for MPC while using standard tools like CasADi and IPOPT. It is also an example use case and a proof of concept for formulating MPC constraints that are inspired by fuzzy logic. For future work, the second interpretation might prove fruitful: instead of shifting the edges of the polygon like in \refEquation{eq:shifting_vertecies}, one could formulate the condition $b_{s,r}' - \matvar{a}_{s,r}' \matvar{p} + d_{\text{min}}(\alpha_{\text{robot}} - \alpha_{\text{edge}}) \geq 0$ in \refEquation{eq:merging_halfspaces_and_polygons_logically} for describing whether a robot collides with an edge of an polygon. The minimum distance $d_{\text{min}}$ would depend on the heading of the robot relative to the orientation of the edge. This way different footprints of the robot could be considered, instead of the often very conservative assumption of a circular footprint. The concept can also be used in different applications for converting verbally formulated constraints into optimization constraints.







\bibliographystyle{IEEEtran}
\bibliography{sources.bib}

\end{document}